\documentclass[authoryear,preprint,review,12pt]{elsarticle}

\usepackage{amssymb}
\usepackage{amsmath}

\usepackage{hyperref}
\usepackage{subfigure}

\usepackage{amsmath}

\usepackage{makecell}   
\usepackage{array}      

\usepackage{algorithm}
\usepackage{algorithmic}

\usepackage{amssymb}
\usepackage{amsthm}
\usepackage{enumerate}
\usepackage{multirow}
\usepackage{xcolor}
\usepackage{framed}
\usepackage{enumitem}

\newtheorem{theorem}{Theorem}

\newtheorem{definition}{Definition}
\newtheorem{assumption}{Assumption}
\newtheorem{remark}{Remark}

\usepackage{newfloat}
\usepackage{listings}
\usepackage{booktabs}
\usepackage{url}
\journal{Arxiv}

\begin{document}
\begin{frontmatter}
\title{On the Stability and Generalization of First-order Bilevel Minimax Optimization}

\author[a]{Xuelin Zhang}
\ead{zhangxuelin@webmail.hzau.edu.cn}

\author[b]{Peipei Yuan}
\ead{yuanpp822@163.com}

\affiliation[a]{organization={College of Informatics},
addressline={Huazhong Agricultural University}, 
city={Wuhan},
postcode={430070}, 
country={China}}
\affiliation[b]{organization={School of Artificial Intelligence},
addressline={Jianghan University}, 
city={Wuhan},
postcode={430056}, 
country={China}}

\begin{abstract}
Bilevel optimization and bilevel minimax optimization have recently emerged as unifying frameworks for a range of machine-learning tasks, including hyperparameter optimization and reinforcement learning. The existing literature focuses on empirical efficiency and convergence guarantees, leaving a critical theoretical gap in understanding how well these algorithms generalize. 
To bridge this gap, we provide the first systematic generalization analysis for first-order gradient-based bilevel minimax solvers with lower-level minimax problems. Specifically, by leveraging algorithmic stability arguments, we derive fine-grained generalization bounds for three representative algorithms, including single-timescale stochastic gradient descent-ascent, and two variants of two-timescale stochastic gradient descent-ascent. Our results reveal a precise trade-off among algorithmic stability, generalization gaps, and practical settings. Furthermore, extensive empirical evaluations corroborate our theoretical insights on realistic optimization tasks with bilevel minimax structures.

\end{abstract}

\begin{keyword}

Bilevel minimax optimization \sep generalization error \sep learning theory

\end{keyword}

\end{frontmatter}

\section{Introduction}

Stochastic bilevel optimization \citep{BrackenM73a,hyper_convergence1, DBLP:conf/nips/BaoWLZZ21} constitutes the optimization problem with hierarchical structures formulated as follows:
\begin{equation}\label{bilevel_problem}
\begin{aligned}
\min_{x \in \mathbb{R}^{d_x}} R(x) &=F\left(x, y^*(x)\right)
:=\mathbb{E}_{\xi}\left[f\left(x, y^*(x) ; \xi\right)\right] \\
\text { s.t. } y^*(x) &\in\arg \min _{y \in \mathbb{R}^{d_y}}\left\{G(x, y):=\mathbb{E}_\zeta[g(x, y; \zeta)]\right\},
\end{aligned}
\end{equation}
where $d_x, d_y \in \mathbb{N}^+$, the outer objective function $f$ and the inner objective function $g$ are both continuous and differentiable, $\xi,\zeta$ are samples drawn from the validation (meta) set and training set, respectively.
Stemming from \citep{BrackenM73a}, the bilevel optimization in Eq.\eqref{bilevel_problem} with minimized lower problem has been extensively utilized for many machine learning fields, e.g., hyper-parameter optimization \citep{DBLP:conf/icml/FranceschiDFP17,franceschi2018bilevel,DBLP:journals/jmlr/OkunoTKW21}, multi-task learning \citep{hu2022multi,ye2024first}, reinforcement learning \citep{DBLP:conf/nips/TschiatschekGHD19,zheng2024safe,shen2025principled}, and few-shot learning \citep{shao2022improving,qin2023bi}. 
However, in specific scenarios such as multi-objective scenarios \citep{giovannelli2024bilevel}, adversary learning, and pessimistic bilevel optimization, the bilevel scheme is further formulated as a minimax lower problem, known as Bilevel Minimax Optimization (BMO). Among these algorithms, (first-order) gradient-based methods \citep{liu2022bome} are preferred for their efficiency and scalability.

In the BMO setting, it's assumed that the lower problem reaches a \emph{saddle point} \citep{yang2020global,chen2022single,huang2025efficiently}, which is most favorable to the upper-level’s objective $F$ \citep{giovannelli2024bilevel}. The mathematical representation for BMO is present as follows,
\begin{equation}\label{bmo_problem}
\begin{aligned}
&\min_{x \in \mathbb{R}^{d_x}} \mathbb{E}_{\xi}\left[f\left(x, y^*(x), z^*(x) ; \xi\right)\right] \\
&\text { s.t. } y^*(x), z^*(x) =  \underset{y \in \mathbb{R}^{d_y}}{\arg \min} ~ \underset{z \in \mathbb{R}^{d_z}}{\arg\max} ~ \mathbb{E}_\zeta[g(x, y,z; \zeta)],
\end{aligned}
\end{equation}
where variable $y$ (or $z$) aims to minimize (or maximize) the lower object with hyperparameter $x$ learned from the upper problem. This complex three-level structure presents unique challenges for algorithm design and statistical analysis. Another scheme, named minimax bilevel optimization (MBO), is formulated with a minimax upper problem and a minimization lower problem.

\begin{table*}[!t]
\centering
\tabcolsep=0.014\linewidth
\resizebox{\textwidth}{!}
{\begin{tabular}{c c c c c }
\toprule 
References  & Target & Analysis Technique & \makecell{Objective Structure\\ Levels (Upper; Lower)} & Algorithm Range \\\midrule
\cite{li2021high} 	&Generalization 	& \makecell{Uniform Stability\\ \& Agument-Stability} 	&1-level (Min-Max)  	&  \makecell{SGDA, PPM, ESP,\\ EG, OGDA} \\
\cite{lei2021stability} 	&Generalization	&  \makecell{Uniform-stability\\ \& Agument-stability} 	&1-level (Min-Max) 	&SGDA, AGDA\\
\cite{farnia2021train}	&Generalization	&Uniformly Stability	&1-level (Min-Max) 	&SGDA\\
\cite{sharma2022federated} 	& Convergence	 & $\epsilon$-stationary  	&1-level (Min-Max) 	&Local SGDA\\
\cite{ozdaglar2022good} 	& Generalization 	&  $\epsilon$-stable 	&1-level (Min-Max) 	& GDA, GDMax\\
\midrule
\cite{DBLP:conf/nips/BaoWLZZ21} & Generalization & Uniform Stability &Bilevel (Min; Min) & UD, TSGD \\
\cite{farid2021generalization} & Generalization & \makecell{PAC-Bayes\\ \& Uniform Stability} &Bilevel (Min; Min) & SGD, GD \\ 
\cite{chen2021generalization} & Generalization & Mutual Information &Bilevel (Min; Min) & UD, TSGD \\ 
\cite{shu2023learning} & Generalization & SRM \& Gaussian Complexity &Bilevel (Min; Min) & SSGD, TSGD \\  
\cite{chen2024exploring}& Generalization & Uniform Stability &Bilevel (Min; Min) &  TSGD \\ 
\cite{wang2024lower}& Lower Bounds & Uniform Stability &Bilevel (Min; Min) & UD, TSGD \\ 
\cite{wang2024stability}& Generalization & Uniform Stability &Bilevel (Min; Min) & TSGD \\   
\cite{pmlr-v235-pan24e}& Generalization  & Uniform Stability &Bilevel (Min; Min) & \makecell{SGD, TSGD,\\ K-level Optimization}\\ 
\midrule
\cite{hu2021stochastic} 	&Convergence	&$\epsilon$-stationary 	&Bilevel (Minimax; Min)	&SSGDA\\
\cite{hu2022multi}	&Convergence	& $\epsilon$-stationary 	&Bilevel (Minimax; Min)	&SGA, SGD\\ 
\cite{gu2023min}	& \makecell{Convergence \\ \& Generalization}	& \makecell{$\epsilon$-(near)-stationary \\ \&Radmacher Complexity} 	&Bilevel (Minimax; Min)	&(Projected) SGD\\
\cite{yang2024first}	&Convergence	&$\epsilon$-stationary 	& Bilevel (Minimax; Min) 	&SSGDA, TSGDA-1\\ 
\midrule
Ours& Upper Bounds & On-average Argument Stability & Bilevel (Min; Minimax) & \makecell{SSGDA, TSGDA-1,\\ TSGDA-2}\\
\bottomrule 
\end{tabular}}
\caption{Related theoretical research on the generalization or convergence guarantees for minimax optimization (upper panel), bilevel optimization (middle panel) and minimax bilevel optimization (lower panel). Notably, SRM, AGDA and GDMax stand for structure risk minimization, alternating gradient descent ascent and gradient descent-max, respectively.}\label{t1_highlight}
\end{table*}

Recently, several optimization algorithms have been developed by introducing the minimax problem into the bilevel scheme.
A straightforward modeling strategy is equipping the deep adversarial training \citep{madry2017towards}, generative adversarial models (GAN) \citep{arjovsky2017wasserstein, yao2024manifold}, and distributionally robust optimization \citep{blanchet2024distributionally} with a bilevel learning scheme, which has been widely applied in image processing and recovery \citep{semibgan,jiang2023bilevel,bilevelgan}, demand response system \citep{kis2021optimistic}, and contextual optimization \citep{bucarey2024decision}. 
Another branch mainly focuses on the \emph{optimistic bilevel optimization}. \citet{wiesemann2013pessimistic} propose an iterative discretization algorithm for solving the pessimistic bilevel optimization problem without the convexity assumption. 
\citet{ye1995optimality} develop approximate KKT conditions via variational analysis and optimality theory for optimistic bilevel optimization problems under non-smoothness and multi-objectivity.
\citet{benchouk2021scholtes} extend the inexact Scholtes relaxation to the optimistic bilevel scheme by dynamically adjusting relaxation errors, addressing robust equilibrium problems in decentralized systems.
\citet{guan2025adaprox} propose a value-function-based reformulation with non-unique lower-level solutions via Lagrangian augmentation, achieving convergence to stationary points.

These algorithms can be roughly categorized into three branches: single-timescale stochastic gradient descent-ascent (SSGDA), two-timescale stochastic gradient descent-ascent with one inner loop (TSGDA-1), and two-timescale stochastic gradient descent-ascent with two inner loops (TSGDA-2).
Nevertheless, in practice, their hyperparameters, including training data sizes, step sizes, and maximum iterations, are empirically selected to achieve better practice generalization performance on limited testing sets. In contrast, the theoretical understanding of hyperparameters' effects on algorithmic generalization with respect to the distribution of testing sets is less well investigated for BMO.
Existing studies on BMO have primarily focused on the computational aspects and convergence behaviors of these methods, leaving a critical gap in our understanding of their generalization performance and the relevance to practical settings. 

\subsection{Related Work}

\emph{Analysis Tools and algorithmic stability.}
Traditional generalization analysis relies on hypothesis space-dependent tools (e.g., the covering numbers, Rademacher complexity, Gaussian complexity, U-statistics \citep{kaariainen2005generalization}, and fractal dimensions \citep{tan2024limitations}). In contrast, these tools largely overlook optimization mechanisms and practical parameter settings, limiting their applicability to complex iterative methods.
Stability-driven generalization theory originated in the 1970s \citep{Ann.Stat/RogersW1978} and has evolved through multiple pivotal formulations: uniform stability \citep{deng2023stability,sun2022stability}, uniform argument stability \citep{liu2017algorithmic}, on-average stability \citep{deng2021toward}, and \emph{on-average argument stability} \citep{DBLP:conf/icml/LeiY20}. 
Notably, the on-average argument stability quantifies the expected parameter sensitivity under distribution shifts, and highlights the tradeoffs between generalization and optimization via adaptive stability constraints. This motivates us to select it for assessing the generalization performance and its relevance to practical optimization settings.

\emph{Generalization theory for minimax optimization.}
Recently, \citet{lei2021stability} study a comprehensive generalization analysis of stochastic gradient methods for minimax problems of stochastic gradient descent ascent (SGDA) under both convex-concave and nonconvex-nonconcave cases. \citet{farnia2021train} present the impact of the optimization algorithm on the generalization performance of the trained minimax model, and uses algorithmic stability to analyze the generalization properties of SGDA and proximal point method (PPM) algorithms. \citet{sharma2022federated} investigate a novel and tighter analysis on the order-optimal sample complexity and linear speedup with the number of clients for Local SGDA. \citet{li2021high} provide the high probability generalization bounds for several popular gradient-based optimization algorithms, including SGDA, PPM, empirical saddle point (ESP), extra-gradient (EG), and optimistic gradient descent ascent (OGDA). Furthermore, \citet{ozdaglar2022good} explore generalization guarantees for minimax optimization problems via the primal gap in nonconvex-concave settings. However, these results for single-level minimax problems do not apply to BMO scenarios with a bilevel scheme.

\emph{Generalization guarantees for bilevel optimization.}
\citet{DBLP:conf/nips/BaoWLZZ21} establish the upper bounds of generalization errors via uniform stability for bilevel hyperparameter optimization with two-timescale stochastic gradient descent (TSGD), compared to unrolled differentiation algorithms. Correspondingly, \citet{wang2024lower} also derive the lower bounds for hyperparameter optimization algorithms with the uniform stability technique. \citet{farid2021generalization} introduce a PAC-Bayesian task-decomposition framework for bilevel algorithms under non-i.i.d. scenarios. 
\citet{zhang2024sbo} derive tight stability-generalization linkages for stochastic bilevel optimization, achieving generalization bounds for single-timescale SGD (SSGD) and TSGD under various convexity scenarios. \citet{chen2024exploring} give the generalization bounds for approximate implicit differentiation (AID)-based bilevel algorithms. \citet{pmlr-v235-pan24e} provide stability guarantees for momentum-based \(K\)-level optimizers, establishing generalization bounds in nonconvex-nonstrongly convex regimes. Nevertheless, none of these considered the minimax problem as the lower problem, which leaves the generalization analysis of BMO still largely underinvestigated.

\emph{Learning Theory for MBO. }
For the multi-block MBO problems under the nonconvex-concave minimax setting, \citet{hu2022multi} present two single-loop randomized stochastic algorithms, which can converge to $\varepsilon$-stationary point.
\citep{hu2021stochastic} develops a single-loop single-timescale stochastic momentum algorithm, reaching the $\mathcal{O}(\epsilon^{-4})$ oracle complexity (via $\epsilon$-stationary) that matches the optimal rate for stochastic nonconvex optimization under a general unbiased oracle model.
\cite{gu2023min} propose a single-loop two-timescale algorithm for multi-task robust learning, which attains convergence and generalization guarantees via Rademacher complexity on worst-case tasks.
To reduce the computational cost and memory consumption in multi-module minimax bilevel optimization, \citet{yang2024first} propose a fully first-order single-loop algorithm with fine-grained analysis on algorithmic generalization. 
Except for \citep{gu2023min}, these works mainly focus on computational complexity and convergence analysis. Moreover, the Rademacher complexity in \citep{gu2023min} does not well illustrate the roles of optimization and experimental settings on algorithmic generalization.

Despite impressive progress in generalization analysis for minimax optimization, bilevel optimization and MBO, there is scarce theoretical work on BMO algorithms. Instead of upper minimax solvers for MBO, the inherent minimax lower problem of BMO introduces a complex, three-level-like structure (min; max-min), which poses significant challenges for the direct theoretical analysis of gradient-based techniques developed for minimax optimization and bilevel optimization. 
Toward this end, the following fundamental problem remains largely open: 

\begin{framed}
\textbf{\emph{How do BMO algorithms theoretically perform in generalization under various experimental settings?}}
\end{framed}

\subsection{Contributions}

To address the aforementioned challenges, this paper conducts a systematic generalization analysis of representative first-order gradient-based BMO algorithms.
Specifically, we first establish the connection between the \emph{on-average argument stability} and generalization errors of the meta learner on BMO under mild conditions. Subsequently, we conduct a detailed analysis of generalization error tailored to each algorithm's unique characteristics. Taking SSGDA as an example, we analyze its stability and then derive an upper bound on its generalization error that depends on the step-size schedule. Additionally, we establish upper bounds on its optimization error and excess risk, achieving the desirable $\mathcal{O}(1/m_1)$ order. These theoretical results, to some extent, elucidate the impact of parameter selection on the model's generalization performance in practical applications, providing valuable guidance for optimizing these parameters.

The main contributions are summarized below:
\begin{itemize}
\item Establishing the unified generalization framework for gradient-based bilevel minimax optimization algorithms. By leveraging on-average argument stability, it derives fine-grained generalization bounds for three representative first-order gradient-based methods, including SSGDA, TSGDA-1 and TSGDA-2, without convexity assumptions. The bounds explicitly quantify the trade-offs between algorithmic stability, generalization performance, and practical optimization settings under mild assumptions. Empirical results also validate our theoretical findings.

\item Providing sharp excess risk and optimization error guarantees for the SSGDA algorithm under convex conditions. The optimization error reaches $\mathcal{O}\left( \frac{V}{\eta^\prime T} +\eta^\prime l^2 \right)$ and the excess risk reaches $\mathcal{O}\left(1/\sqrt{m_1}\right)$. This insight highlights the increased complexity of BMO resulting from error accumulation in its hierarchical min-max-min structure.

\end{itemize}

To better highlight the contributions, Table \ref{t1_highlight} summarizes the comparisons with some latest related works on generalization analysis for minimax, bilevel optimization and MBO algorithms. To the best of our knowledge, the generalization analysis of the minimax problem at the lower level under bilevel schemes has not been previously discussed.

\section{Problem Formulation} 

Given distributions $\mathbb{D}_1$, $\mathbb{D}_2$, we get the validation (meta) set $\mathcal{D}_{m_1} := \{\xi_i\}_{i=1}^{m_1} \sim \mathbb{D}_1^{m_1}$ and
the training set $\mathcal{D}_{m_2} := \{\zeta_i\}_{i=1}^{m_2} \sim \mathbb{D}_2^{m_2}$
by independent sampling, where $m_1$ and $m_2$ are the sample sizes. This paper focuses on the outer-level population risk w.r.t $\mathbb{D}_1$ and empirical risk w.r.t $\mathcal{D}_{m_1}$. 

Denote $x \in \mathcal{X}$ as the upper variable, $ y \in \mathcal{Y}$ and $z\in\mathcal{Z}$ are the inner variables for the minimax problem.
Now we introduce the upper-level population risk w.r.t $\mathbb{D}_1$ and the  empirical risk w.r.t $x$ $\mathcal{D}_{m_1}$ defined respectively as
\begin{equation}\label{population_risk}
\mathcal{R}\left(x,y,z\right)=\mathbb{E}_{\xi \sim \mathbb{D}_1}[f(x, y(x),z(x); \xi)],
\end{equation}
and
\begin{equation}\label{empirical_risk}
\mathcal{R}_{\mathcal{D}_{m_1}}\left(x,y,z\right)=\frac{1}{m_1} \sum_{i=1}^{m_1}\left[f\left(x, y(x),z(x); \xi_i\right)\right],
\end{equation}
where $f$ is the objective function at the upper level. $y(x)$ and $z(x)$ are the inner model parameters given the outer model parameter $x$ (also see Eq.\eqref{bmo_problem}).

Let $(x, y(x), z(x))$ in Eq.\eqref{bmo_problem} be estimated by a stochastic algorithm $A$ with data $\mathcal{D}_{m_1}, \mathcal{D}_{m_2}$, i.e. $A\left(\mathcal{D}_{m_1}, \mathcal{D}_{m_2}\right)$.  
Similar to the previous works \citep{DBLP:conf/nips/BaoWLZZ21,hoffer2017train,keskar2017on} for evaluating the approximated searching of hyperparameters, we define
\begin{equation}\label{generalization_gap}
\mathbb{E}\left[\mathcal{R}\left(A\left(\mathcal{D}_{m_1}, \mathcal{D}_{m_2}\right)\right)-\mathcal{R}_{\mathcal{D}_{m_1}}\left(A\left(\mathcal{D}_{m_1}, \mathcal{D}_{m_2}\right)\right)\right]
\end{equation}
in the upper (outer) level as the generalization gap of $A$, which measures the difference between the population risk $\mathcal{R}(A)$ in Eq.\eqref{population_risk} and the empirical risk $\mathcal{R}_{\mathcal{D}_{m_1}}(A)$ in Eq.\eqref{empirical_risk}.

\subsection{Learning Algorithms}

To characterize the generalization behaviors of the general gradient-based bilevel minimax models for solving \eqref{bmo_problem}, we consider three typical optimization algorithms, including the SSGDA in Algorithm 1, TSGDA-1 and TSGDA-2 (see Algorithms 2 and 3 in the Appendix). SSGDA differs slightly from TSGDA-1 by setting the inner iteration $K=1$.

With the bilevel scheme, the computed gradients $\nabla_x$ may contain the second-order information or simplified first-order information. Due to space limitations, the algorithms of TSGDA-1 and TSGDA-2 are left in \emph{Appendices D and E}, respectively.

\subsection{Excess Risk and Error Decomposition}

For simplicity, denote $\mathbb{E}_{A, \mathcal{D}_{m_1}, \mathcal{D}_{m_2}}$ by $\mathbb{E}$.
Motivated by previous works \citep{DBLP:conf/nips/BaoWLZZ21,pmlr-v235-pan24e}, the algorithmic capability of the meta learner for searching hyperparameters is evaluated by denoting 
\begin{equation}\label{risk_1}
\begin{aligned}
\mathcal{R}^*\left(x,y,z\right)
=\mathcal{R}\left(x^*, y,z\right) 
:= \inf_{x \in \mathcal{C}_{x}} \mathcal{R}\left(x, y,z\right),
\end{aligned}
\end{equation}
and
\begin{equation}\label{risk_2}
\begin{aligned}
\widehat{\mathcal{R}}_{\mathcal{D}_{m_1}} \left(A\left(x,y,z\right)\right)
&:=\mathcal{R}_{\mathcal{D}_{m_1}} \left(A\left(\widehat{x},y,z\right)\right)\\
&:=\inf_{x \in \mathcal{C}_{x}} \mathcal{R}_{\mathcal{D}_{m_1}}\left(x, y,z\right).
\end{aligned}
\end{equation}

The \emph{excess risk} \citep{bousquet2002stability} at the upper level is formulated by
\begin{equation}
\begin{aligned}
&\mathbb{E}\left[\mathcal{R} \left(A\left(\mathcal{D}_{m_1}, \mathcal{D}_{m_2}\right)\right) -  \mathcal{R}^* \left(A\left(\mathcal{D}_{m_1}, \mathcal{D}_{m_2}\right)\right)\right] \\
\leq &\mathbb{E}\left[\mathcal{R} \left(A\left(\mathcal{D}_{m_1}, \mathcal{D}_{m_2}\right)\right) -  \mathcal{R}_{\mathcal{D}_{m_1}} \left(A\left(\mathcal{D}_{m_1}, \mathcal{D}_{m_2}\right)\right)\right] \\
+ &\mathbb{E}\left[\mathcal{R}_{\mathcal{D}_{m_1}} \left(A\left(\mathcal{D}_{m_1}, \mathcal{D}_{m_2}\right)\right) -  \widehat{\mathcal{R}}_{\mathcal{D}_{m_1}} \left(A\left(\mathcal{D}_{m_1}, \mathcal{D}_{m_2}\right)\right)\right],
\end{aligned}
\end{equation}
since $\widehat{\mathcal{R}}_{\mathcal{D}_{m_1}}\left(x^*,y,z\right)\leq \mathcal{R}_{\mathcal{D}_{m_1}}\left(x^{*},y,z\right)$ holds by definitions in Eq.\ref{risk_1} and \ref{risk_2}.
The first counterpart is defined in Eq.\eqref{generalization_gap} as the generalization gap of algorithm $A$. 

Moreover, the second term 
\begin{equation*}
\mathbb{E}\left[\mathcal{R}_{\mathcal{D}_{m_1}} \left(A\left(\mathcal{D}_{m_1}, \mathcal{D}_{m_2}\right)\right) -  \widehat{\mathcal{R}}_{\mathcal{D}_{m_1}} \left(A\left(\mathcal{D}_{m_1}, \mathcal{D}_{m_2}\right)\right)\right]
\end{equation*}
is denoted as the \emph{optimization error}, which quantifies the algorithmic efficiency in minimizing empirical risk.

\subsection{Preliminaries}

Before introducing our results,  we first recall some necessary assumptions and the definitions of \emph{on-average argument stability} \citep{DBLP:conf/icml/LeiY20}, which have been extensively used in convergence analysis \citep{hyper_convergence1,ghadimi2018approximation,liu2022bome} and generalization analysis \citep{DBLP:conf/icml/HardtRS16,lin2017distributed,lei2021stability,lei2023generalization} for single-level and bilevel optimization.

\begin{assumption}\label{gen_assumtion_continuity}
(Joint Continuity \citep{hyper_convergence1,liu2022bome}). Assume that $f$ is jointly $l_f$-Lipschitz continuous w.r.t. $x, y,z$, there holds
\begin{equation*}
\begin{aligned}
\left| f(x,y,z;\xi)-f(x^\prime,y^\prime,z^\prime;\xi)\right| \leq l \left\|\begin{array}{c}
x-x^{\prime} \\
y-y^{\prime}\\
z-z^{\prime}\\
\end{array}\right\|_2,
\end{aligned}
\end{equation*}
for any $(x,y,z), (x^\prime,y^\prime,z^\prime)\in \mathcal{X} \times \mathcal{Y} \times \mathcal{Z}$ and $\xi\sim \mathbb{D}_1$. Similarly, let $g$ also be jointly $l_g$-Lipschitz continuous.
\end{assumption}

\begin{assumption}\label{gen_assumtion_smooth}
(Joint Smoothness \citep{lei2021stability}). Function $f$ is jointly $L_f$-smooth w.r.t. $x, y$ and $z$, there holds
\begin{equation*}
\begin{aligned}
&\|\nabla f(x,y,z;\xi)-\nabla f(x^\prime,y^\prime,z^\prime;\xi)\|_2  \leq L \left\|\begin{array}{c}
x-x^{\prime} \\
y-y^{\prime}\\
z-z^{\prime}\\
\end{array}\right\|_2,
\end{aligned}
\end{equation*}
for any $(x,y,z), (x^\prime,y^\prime,z^\prime)\in \mathcal{X} \times \mathcal{Y} \times \mathcal{Z}$ and $\xi\sim \mathbb{D}_1$. Similarly, let $g$ also be jointly $L_g$-smooth.
\end{assumption}

\begin{definition}\label{definition_4}
(H$\ddot{o}$lder Continuity). Let $\tau>0, \alpha \in[0,1]$. Gradient $\nabla f$ is $(\alpha,\tau)$-H$\ddot{o}$lder continuous over $\mathbb{\mathbb{R}}^{d_x} \times \mathbb{\mathbb{R}}^{d_y}\times \mathbb{\mathbb{R}}^{d_z}$, if there holds
\begin{equation*}
\|\nabla f(x,y,z;\xi)-\nabla f( x^{\prime},y^{\prime},z^{\prime}; \xi)\|_2 \leq \tau \left\|\begin{array}{l}
x-x^{\prime} \\
y-y^{\prime} \\
z-z^{\prime} \\
\end{array}\right\|_2^\alpha,
\end{equation*}
for all $(x,y,z), (x^\prime,y^\prime,z^\prime)\in \mathcal{X} \times \mathcal{Y} \times \mathcal{Z}$ and $\xi\sim \mathbb{D}_1$.
\end{definition}

\begin{remark}
H$\ddot{o}$lder continuity is weaker than Lipschitz continuity and smoothness \citep{DBLP:conf/icml/LeiY20,nesterov2015universal}. As parameterized in Definition \ref{definition_4}, $\alpha=1$ implies the smoothness, while $\alpha=0$ corresponds to the Lipschitz continuity (allowing non-differentiability) \citep{DBLP:conf/icml/LeiY20}. This assumption encompasses objective functions such as the mean absolute function, hinge function, and their variants \citep{DBLP:conf/icml/LeiY20,steinwart2008support}.
\end{remark}

\begin{definition}\label{definition_5}
(On-average Argument Stability). Let $\mathcal{D}_{m_1}=\left\{\xi_1, \ldots, \xi_{m_1}\right\}$ and $\widetilde{D}_{m_1}=\left\{\tilde{\xi}_1, \ldots, \tilde{\xi}_{m_1}\right\}$ be two sets drawn independently from distribution $ \mathbb{D}_1^{m_1}$. For any $i=1, \ldots, m_1$, define $D^{(i)}=$ $\left\{\xi_1, \ldots, \xi_{i-1}, \tilde{\xi}_i, \xi_{i+1}, \ldots, \xi_{m_1}\right\}$. Denote the $\mathbb{E}$ as the expectation of $\mathbb{E}_{\mathcal{D}_{m_1}, \mathcal{D}_{m_2},\widetilde{D}_{m_1}, A}$.
We say a randomized algorithm $A$ is $l_1(\beta)$ on-average argument stable if
\begin{equation*}
\mathbb{E}\left[\frac{1}{m_1} \sum_{i=1}^{m_1}\left\|A(\mathcal{D}_{m_1},\mathcal{D}_{m_2})-A\left(\mathcal{D}_{m_1}^{(i)},\mathcal{D}_{m_2}\right)\right\|_2\right] \leq \beta,
\end{equation*}
and $l_2(\beta^2)$ on-average argument stable if
\begin{equation*} 
\begin{aligned}
\mathbb{E}\left[\frac{1}{m_1} \sum_{i=1}^{m_1}\left\|A(\mathcal{D}_{m_1},\mathcal{D}_{m_2})-A\left(\mathcal{D}_{m_1}^{(i)},\mathcal{D}_{m_2}\right)\right\|_2^2\right] \leq \beta^2.
\end{aligned}
\end{equation*}
\end{definition}

\begin{remark}
The on-average argument stability quantifies the expected parameter sensitivity to single-sample validation set perturbations. This differs fundamentally from uniform stability \citep{DBLP:conf/nips/BaoWLZZ21}, which measures prediction error drift and requires loss boundedness.
\end{remark}

Based on the above conditions, we can further derive the upper bounds of the generalization gap for several stochastic bilevel minimax optimization algorithms, including SSGDA, TSGDA-1 and TSGDA-2.

\section{Theoretical Assessments}

For stochastic BMO problems \eqref{bmo_problem}, this section presents a generalization analysis of three representative algorithms within a bilevel optimization framework.
For simplicity, denote $\eta^t$, $T$ (or $\gamma^k$, $K$ or $\gamma^q$, $Q$) as the stepsize at $t$-th (or $k$-th or $q$-th) iteration and maximum iterations for the upper problem (or lower problems).
All proofs of theoretical results can be found in \emph{Appendices B to E}.

\subsection{Connection between Stability and Generalization}

This section states that the generalization gap of BMO's meta learner \eqref{bmo_problem} can be bounded by the on-average argument stability. 

\begin{theorem}\label{theorem1}
(\textrm{I}) If algorithm $A$ is $l_1(\beta)$ on-average argument stable in expectation and the outer-level function $f$ is $L_f$-Lipschitz continuous w.r.t. $(x,y) \in \mathbb{R}^{d_1} \times \mathbb{R}^{d_2}$, denote $\mathbb{E}$ as $\mathbb{E}_{A, \mathcal{D}_{m_1}, \mathcal{D}_{m_2}}$, there holds 
\begin{equation*}
\begin{aligned}
&|\mathbb{E}\left[\mathcal{R}\left(A\left(\mathcal{D}_{m_1}, \mathcal{D}_{m_2}\right)\right)-\mathcal{R}_{\mathcal{D}_{m_1}}\left(A\left(\mathcal{D}_{m_1}, \mathcal{D}_{m_2}\right)\right)\right]| 
\leq \ell_f \beta.
\end{aligned}
\end{equation*}

(\textrm{II}) If algorithm $A$ is $l_2(\beta^2)$ on-average argument stable in expectation and f is nonnegative and $L_f$-smooth w.r.t. $(x,y) \in \mathbb{R}^{d_1} \times \mathbb{R}^{d_2}$, denote $\mathbb{E}$ as $\mathbb{E}_{A, \mathcal{D}_{m_1}, \mathcal{D}_{m_2}}$, then 
\begin{equation*}
\centering
\begin{aligned}\label{l2_stability}
& \mathbb{E}\left[\mathcal{R}\left(A\left(\mathcal{D}_{m_1}, \mathcal{D}_{m_2}\right)\right)-\mathcal{R}_{\mathcal{D}_{m_1}}\left(A\left(\mathcal{D}_{m_1}, \mathcal{D}_{m_2}\right)\right)\right] \\
\leq  &  \frac{L_f}{\gamma} \mathbb{E}\left[\mathcal{R}_{\mathcal{D}_{m_1}}(A(\mathcal{D}_{m_1},\mathcal{D}_{m_2}))\right] 
+\frac{(L_f+\gamma) \beta^2}{2} ,
\end{aligned}
\end{equation*}
where the constant $\gamma>0$. 

(\textrm{III}) If algorithm $A$ is $l_2(\beta^2)$ on-average argument stable in expectation, f is nonnegative and $(\alpha,\tau)$-H$\ddot{o}$lder continuous w.r.t. $(x,y) \in \mathbb{R}^{d_1} \times \mathbb{R}^{d_2}$ with $\alpha \in [0,1]$, then 
\begin{eqnarray*}
&& \mathbb{E}\left[\mathcal{R}\left(A\left(\mathcal{D}_{m_1}, \mathcal{D}_{m_2}\right)\right)-\mathcal{R}_{\mathcal{D}_{m_1}}\left(A\left(\mathcal{D}_{m_1}, \mathcal{D}_{m_2}\right)\right)\right] \\
&\leq&  \frac{c_{\alpha, \tau}^2}{2 \gamma} \mathbb{E}\left[\mathcal{R}^{\frac{2 \alpha}{1+\alpha}}(A\left(\mathcal{D}_{m_1}, \mathcal{D}_{m_2}\right))\right] + \frac{\gamma}{2} \beta^2,
\end{eqnarray*}
for $\mathcal{D}_{m_1} \sim \mathbb{D}_1^{m_1}$ and $\mathcal{D}_{m_2} \sim \mathbb{D}_2^{m_2}$,
with constant $\gamma>0$.
\end{theorem}

\begin{remark}
Theorem \ref{theorem1} validates the connection between on-average argument stability and the generalization gap. Especially, the smoothness assumption is further relaxed by the H$\ddot{o}$lder continuity in Theorem \ref{theorem1}(III).
\end{remark}

\begin{table*}[!t]
\centering 
\scalebox{1}{ 
\resizebox{\textwidth}{!}{
\begin{tabular}{ccccccccccc}
\toprule
\centering
Algorithms &\centering  Object &\centering Assumptions &\centering Step Sizes&\centering Bounds  \tabularnewline\midrule
\multirow{4}{*}{\centering \makecell{SSGDA \\ (Theorem 4)}}  
& Optimization Error & \makecell{Assumptions 1,2 \& Convexity\\  $\mathbb{E}\left[\|x^0 - \widehat{x}\|\right] \leq V$ }&  --- &$\mathcal{O}\left( \frac{V}{\eta^\prime T} +\eta^\prime l^2 \right)$ \\ \cmidrule{2-5} 
& Excess Risk
& \makecell{Assumptions 1,2 \& Convexity\\  $\mathbb{E}\left[\|x^0 - \widehat{x}\|\right] \leq V$ \& $T\asymp \sqrt{m_1}$} & \makecell{$\eta^{t \prime} \leq \min\{\frac{C_0}{\sqrt{T}}, \frac{C_1}{t}\}$, $C_1 \in (0,1/7) $} &$\mathcal{O}\left(1/\sqrt{m_1}\right)$  \\
\midrule
\multirow{3}{*}{\centering \makecell{SSGDA \\ (Theorem 3)}}
& \multirow{3}{*}{\centering \makecell{Generalization Error}}
& \multirow{3}{*}{\centering \makecell{Assumptions 1, 2}} &  \multirow{3}{*}{\centering \makecell{$ \eta^{t \prime}\leq C_2/(tL)\leq1$, $C_2 \in (0,1/7]$ \\ $ \eta^{t \prime}\leq C_2/(tL)\leq1$, $C_2 = 1/7$ \\$ \eta^{t \prime}\leq C_2/(tL)\leq1$, $C_2 > 1/7$ \\ ~}}  & \multirow{3}{*}{\centering \makecell{$\mathcal{O}\left(T/m_1\right)$ \\ $\mathcal{O}\left(T\ln(T)/m_1\right)$ \\ $\mathcal{O}\left(T^{7C_2}/m_1\right)$}}  \\
& ~
& ~ &  ~ & ~  \\
& ~
& ~ &  ~ & ~  \\
\midrule
\makecell{TSGDA-1 \\ (Theorem 5)} 
& Generalization Error 
& Assumptions 1, 2 &  \makecell{$\eta^t\leq C_3/(tL+L)\leq 1$, $C_3 \in (0,1)$\\ $\gamma_1^k,\gamma_2^k\leq C_4/(kL+L)\leq 1$, $C_4\in (0,1)$ } & $\mathcal{O}\left(\frac{\sqrt{3}^{T} T^{C_5+1} K^{(C_5+1)T+1}}{m_1}\right)$\\
\midrule
\makecell{TSGDA-2 \\ (Theorem 6)}
& Generalization Error 
& Assumptions 1, 2 &  \makecell{$\eta^t \leq C_6/(tL+L)  \leq 1$, $C_6\in(0,1)$\\$\gamma_1^k \leq C_7/(kL+L)\leq 1,$ $C_7\in(0,1)$\\ $\gamma_2^q \leq C_8/(qL+L)\leq 1$, $C_8\in(0,1)$} & $\mathcal{O}\left(\frac{\sqrt{3}^{T} T^{C_6+1}  \max\{K^{(C_9+1)T+1}, Q^{(C_9+1)T+1}\}}{m_1}\right)$\\
\bottomrule
\end{tabular}}
}
\caption{Summary of the theoretical results of this paper. Notice that positive constants $C_5 = \max\{C_3,C_4\}$, $C_9=\max\{C_6,C_7,C_8\}$
$\eta^{t\prime}= \max\{\eta^t,\gamma_1^t,\gamma_2^t\}$;
$L=\max\{L_f, L_g\}$;
$m_1$ is the number of validation samples; 
$T$, $K$, and $Q$ are the total numbers of outer and inner iterations. 
}
\label{t2_summary}
\end{table*}

\subsection{Generalization Analysis for SSGDA}

Let $\eta$ and $\gamma_1$, $\gamma_2$ be the step sizes for updating $x$, $y$ and $z$. According to Theorem \ref{theorem1}, the on-average argument stable metrics in Definition \ref{definition_5} for SSGDA algorithm with $T$ iterations $\left\|A(\mathcal{D}_{m_1},\mathcal{D}_{m_2})-A\left(\mathcal{D}_{m_1}^{(i)},\mathcal{D}_{m_2}\right)\right\|_2$ can be measured by  
\begin{equation*}
\sqrt{\|x^{T}-x^{T \prime\|_2^2} +  \|y^{T}-y^{T \prime}\|_2^2 +   \|z^{T}-z^{T \prime }\|_2^2},
\end{equation*}
where $x^{T\prime}$ implies variable $x$ training on the corrupted sample $\tilde{\xi}_i$ w.r.t. stability definition at $T$-th outer iteration.

Now we state the upper bounds of on-average argument stability for SSGDA in Algorithm 1 in the Appendix.

\begin{theorem}\label{stability_1}
Let Assumptions \ref{gen_assumtion_continuity}, \ref{gen_assumtion_smooth} be satisfied, and let $A$ be the SSGDA algorithm with $T$ iterations ($T>1$). Denote positive constants $c_0,c_1>0$, $L=\max\{L_f,L_g\}$. Let non-increasing step sizes satisfy $\eta^{t \prime} =\max\{\eta^t, \gamma_1^t,\gamma_2^t\}$, $ \eta^{t \prime}\leq \frac{c_1}{tL}\leq1$.
When $c_1=1/7$, there holds $\beta \leq \mathcal{O}\left(\frac{T \ln (T)}.{m_1}\right)$.
Otherwise, there holds $\beta \leq \mathcal{O}\left(\frac{\max\{T,T^{7c_1}\}}{m_1}\right).$
\end{theorem}

Utilizing Theorem \ref{theorem1} and Theorem \ref{stability_1}, we obtain the following conclusions on generalization bounds for SSGDA.
\begin{theorem} \label{gen_ssgda}
Under the same conditions and settings on step sizes as in Lemma \ref{stability_1}, the generalization error of the SSGDA algorithm satisfies
\begin{equation*}
\begin{aligned}
&|\mathbb{E}\left[\mathcal{R}\left(A\left(\mathcal{D}_{m_1}, \mathcal{D}_{m_2}\right)\right)-\mathcal{R}_{\mathcal{D}_{m_1}}\left(A\left(\mathcal{D}_{m_1}, \mathcal{D}_{m_2}\right)\right)\right]| \\
\leq &\begin{cases}\mathcal{O}\left(\frac{T}{m_1}\right), & \text { if } 0<c_1<1/7, \\ \mathcal{O}\left(\frac{T \ln (T)}{m_1}\right), & \text { if } c_1=1/7, \\ \mathcal{O}\left(\frac{T^{7c_1}}{m_1}\right), & \text { if } c_1>1/7. \end{cases}
\end{aligned}
\end{equation*}

\end{theorem}

\begin{remark}
The results from Theorem \ref{gen_ssgda} indicate that the SSGDA algorithm exhibits a favorable generalization order without relying on the conditions of convexity or concavity. Specifically, this further substantiates that experimental configurations, such as the step sizes and the number of iterations, significantly influence the algorithmic generalization capabilities.
\end{remark}

\begin{theorem}\label{theorem_error_opt} (Optimization error and excess risk)
Let Assumptions \ref{gen_assumtion_continuity}, \ref{gen_assumtion_smooth} be satisfied, and let $A$ be the SSGDA algorithm with $T$ iterations. Denote the step sizes by $\eta^{ \prime} =\max\{\eta, \gamma_1,\gamma_2\}$. Let $\ell = \max\{\ell_f,\ell_g\}$ If function $f$ is convex with respect to $x$ and the domain $\mathcal{X}$ is bounded in $\mathbb{R}^{d_x}$ where $\mathbb{E}\left[\|x^0 - \widehat{x}\|\right] \leq V$, the following result holds
\begin{equation*}
\begin{aligned}
&\mathbb{E}\left[\mathcal{R}_{\mathcal{D}_{m_1}} \left(A\left(\mathcal{D}_{m_1}, \mathcal{D}_{m_2}\right)\right) -  \widehat{\mathcal{R}}_{\mathcal{D}_{m_1}} \left(A\left(\mathcal{D}_{m_1}, \mathcal{D}_{m_2}\right)\right)\right]
\leq \mathcal{O}\left( \frac{V}{\eta^\prime T} +\eta^\prime l^2 \right).
\end{aligned}
\end{equation*}
Moreover, by selecting the number of iterations $T\asymp \sqrt{m_1}$ and the non-increasing step sizes $\eta^{t \prime} =\max\{\eta^t, \gamma_1^t,\gamma_2^t\} \leq \min\{\frac{c}{\sqrt{T}}, \frac{c_1}{t}\}$ where $c>0, 0<c_1<1/7$, we have
\begin{equation*}
\mathbb{E}\left[\mathcal{R} \left(A\left(\mathcal{D}_{m_1}, \mathcal{D}_{m_2}\right)\right) -  \mathcal{R}^* \left(A\left(\mathcal{D}_{m_1}, \mathcal{D}_{m_2}\right)\right)\right]
\leq \mathcal{O}\left(1/\sqrt{m_1}\right).
\end{equation*}
\end{theorem}

\begin{remark}
Theorem \ref{theorem_error_opt} illustrates that by appropriately setting iteration number $T \asymp \sqrt{m_1}$ and step sizes, the excess risk of the SSGDA algorithm reaches the rate of $\mathcal{O}\left(1/\sqrt{m_1}\right)$ under the convex setting. Compared to single-level stochastic gradient descent with $\mathcal{O}\left(1/m_1\right)$ \citep{DBLP:conf/icml/HardtRS16}, the SSGDA algorithm with minimax setting and bilevel scheme requires more iterations and updating loops. The increased generalization error and excess risk may be attributed to the empirically accumulated estimation errors.
\end{remark}

\subsection{Generalization Analysis for TSGDA-1}

Denote $\nabla_y g(x,y,z)$ as the partial derivative of the function $g$ over variable $y$. $y^{t+1,0} = y^{t,K}$ represents the inner parameter $y$ in $K$-th outer loop and $t$-th inner loop. For the TSGDA-1 algorithm with totally $K$ outer iterations and $T$ inner iterations, the argument stability $\left\|A(\mathcal{D}_{m_1},\mathcal{D}_{m_2})-A\left(\mathcal{D}_{m_1}^{(i)},\mathcal{D}_{m_2}\right)\right\|_2$ is measured by 
$\sqrt{\|x^{T}-x^{T \prime\|_2^2} +  \|y^{T}-y^{T,0 \prime}\|_2^2 +  \|z^{T}-z^{T,0 \prime }\|_2^2}$, where there exist single iterative inner loops for updating $y$ and $z$:
$$y^{T,0} = y^{T-1,0} - \sum_{k=0}^{K-1} \eta^t \nabla_y g(x^{T-1},y^{T-1,k},z^{T-1,k}),$$ 
and  
$$z^{T,0} = z^{T-1,0} - \sum_{k=0}^{K-1} \eta^t \nabla_y g(x^{T-1},y^{T-1,k+1},z^{T-1,k}).$$

We now state the upper bounds on the generalization error for the TSGDA algorithm.

\begin{theorem}\label{gen_tsgda1}
Let Assumptions \ref{gen_assumtion_continuity}, \ref{gen_assumtion_smooth} be satisfied and let $A$ be the TSGDA-1 algorithm with $T$ outer iterations and $K$ inner iterations. Denote non-increasing step sizes by $\eta^t\leq\frac{c_2}{(t+1)L}\leq 1$, $L=\max\{L_f,L_g\}$, $\gamma_1^k,\gamma_2^k\leq\frac{c_3}{(k+1)L}\leq 1$ with constants $c_2, c_3\in (0,1)$. Let $c_4 = \max\{c_2,c_3\}$. The stability argument for the TSGDA-1 algorithm satisfies
Then the generalization error for TSGDA-1 satisfies
\begin{equation*}
\begin{aligned}
&|\mathbb{E}\left[\mathcal{R}\left(A\left(\mathcal{D}_{m_1}, \mathcal{D}_{m_2}\right)\right)-\mathcal{R}_{\mathcal{D}_{m_1}}\left(A\left(\mathcal{D}_{m_1}, \mathcal{D}_{m_2}\right)\right)\right]| 
\leq \mathcal{O}\left(\frac{\sqrt{3}^{T} T^{c_2+1} K^{(c_4+1)T+1}}{m_1}\right).
\end{aligned}
\end{equation*}
\end{theorem}

\subsection{Generalization Analysis for TSGDA-2}

For TSGDA-2 algorithm with totally $T$ outer iterations, $K$ inner iterations for updating $y$, and another $Q$ inner iterations for updating $z$, the argument stability $\left\|A(\mathcal{D}_{m_1},\mathcal{D}_{m_2})-A\left(\mathcal{D}_{m_1}^{(i)},\mathcal{D}_{m_2}\right)\right\|_2$ is measured by 
$$\sqrt{\|x^{T}-x^{T \prime\|_2^2} +  \|y^{T}-y^{T,0 \prime}\|_2^2 +  \|z^{T}-z^{T,0 \prime }\|_2^2},$$ 
where there are two individual inner loops 
$$y^{T,0} = y^{T-1,0} - \sum_{k=0}^{K-1} \eta^t \nabla_y g(x^{T},y^{T-1,k},z^{T-1,0}),$$ 
and
$$z^{T,0} = z^{T-1,0} - \sum_{q=0}^{Q-1} \eta^t \nabla_y g(x^{T},y^{T,0},z^{T-1,q}).$$

\begin{theorem} \label{gen_tsgda2}
Let Assumptions \ref{gen_assumtion_continuity}, \ref{gen_assumtion_smooth} be satisfied and let $A$ be the TSGDA-2 algorithm with $T$ outer iterations, $K$ and $Q$ inner iterations. Denote non-increasing step sizes by $\eta^t \leq \frac{c_5}{(t+1) L } \leq 1, \gamma_1^k \leq \frac{c_6}{(k+1)L}\leq 1$ and $\gamma_2^q \leq \frac{c_7}{(q+1)L}\leq 1$ with constants $c_5, c_6,c_7\in(0,1)$ and $L=\max\{L_f,L_g\}$. Let $c_8 = \max\{c_5,c_6,c_7\}$. 
There holds
\begin{equation*}
\begin{aligned}
&|\mathbb{E}\left[\mathcal{R}\left(A\left(\mathcal{D}_{m_1}, \mathcal{D}_{m_2}\right)\right)-\mathcal{R}_{\mathcal{D}_{m_1}}\left(A\left(\mathcal{D}_{m_1}, \mathcal{D}_{m_2}\right)\right)\right]| \\
\leq &\mathcal{O}\left(\frac{\sqrt{3}^{T} T^{c_5+1}  \max\{K^{(c_8+1)T+1}, Q^{(c_8+1)T+1}\}}{m_1}\right).
\end{aligned}
\end{equation*}

\end{theorem}

\begin{remark}
To the best of our knowledge, this is the first work to analyze the generalization performance of general stochastic optimization algorithms for solving BMO with lower-level min-max problems. The exploited algorithmic stability can better illustrate the roles of optimization settings on generalization capability. 
Theorems \ref{gen_tsgda1} and \ref{gen_tsgda2} further illustrate that the cumulative stochastic gradients and corresponding estimation errors brought by increased iterations and additional loops might escalate the algorithmic stability and generalization capability, which is broadly consistent with existing theoretical conclusions on bilevel \citep{DBLP:conf/nips/BaoWLZZ21} and K-level algorithms \citep{pmlr-v235-pan24e}.
\end{remark}

\section{Empirical Validation}

This section empirically validates the generalization performance of the BMO algorithm on real-world data. 
All experiments are implemented in Python on a PC equipped with an Intel(R) Xeon(R) Platinum 8175M CPU and an NVIDIA RTX A6000 GPU. Implemented codes (including \citep{shu2019meta} for bilevel reweighting scheme and \citep{yao2024manifold} for generative models) and data sets (the frames of Charlie Chaplin \footnote{Download from \url{https://github.com/zhigang-yao/MFCGAN}})  are from publicly available sources. Our codes are released at \url{https://github.com/zxlml/bilevel_minimax_generalization}.

\subsection{Experimental Settings}
We practically utilize the TSGDA-1 optimization algorithm (summarized in \emph{Appendix D}) as the solver for the BMO problem, due to its representative scheme and its recovery from SSGDA with an inner loop of $K=1$. 
The task is specified as a general hyperparameter optimization problem \citep{hyper_convergence1} within a bilevel generative adversarial model \citep{semibgan,jiang2023bilevel} for instance reweighting. 
Here, we present empirical findings on the nonlinear interpolation task for Charlie Chaplin's frames using specific GANs \citep{yao2024manifold} with a bilevel reweighted strategy \citep{shu2019meta}. 
The objective function is precisely formulated by the BMO framework in Eq.\ref{bmo_problem}, where $x$ stands for the MLP's weights for instance importance mapping \citep{shu2019meta} at the upper level, $y$ and $z$ are the parameters of the generators and discriminators \citep{yao2024manifold}, respectively.
The average results within inner iterations satisfying $k \in [10,100,150,200,250,300]$ after 10 independent repeats are plotted. The default settings include $m_1=2000, T=50, K=300$ and $\eta$ initialized as $10^{-3}$ with $0.95$ exponential decay rate.

\subsection{Experimental Details and BMO Objectives}

Denote the meta set as $\mathcal{D}_{m_1}$ and training data as $\mathcal{D}_{m_2}$.
Originally, the bilevel weighting scheme of the Meta Weighting Network (MWNet) \cite{shu2019meta} with a minimized lower problem could be formulated by 
\begin{equation*} 
\begin{aligned} 
&\min_{x \in \mathcal{X}} \mathcal{R}_{m_1}(x,y):=\mathbb{E}_{\xi\in\mathcal{D}_{m_1}} f\left(x,y^*(x)\right), ~ \text{ s.t. } ~ y^*(x) \in  \underset{y \in \mathcal{Y}}{\arg\min} ~\mathbb{E}_{\zeta \in \mathcal{D}_{m_2}} g(x,y),
\end{aligned}
\end{equation*}
where $x$ is the parameter of the upper-level MLP with 1-10-10-1 architecture activated by Relu in hidden layers and Sigmoid in the output layer. $g$ is the lower objective function weighted by the MLP, and $y$ is the parameter of the learned function for prediction.

And the selected generative adversarial network for manifold fitting and image processing tasks \cite{yao2024manifold} involves the parameters of two generators as $G_{1}, G_{2}$, two discriminators as $D_{1}, D_{2}$ and a specific module as $F_{\mathcal{M}}$ for manifold fitting, the single-level object is
\begin{equation*}
\begin{aligned}
&G_{1}, G_{2}, F_{\mathcal{M}}, D_{1}, D_{2}
= \underset{F_{\mathcal{M}}, D_{1}, D_{2}}{\arg\min} ~ \underset{G_{1}, G_{2}}{\arg\max} ~ \mathbb{E}_{\zeta \in \mathcal{D}_{m_2}} g \left(F_{\mathcal{M}}, D_{\mathcal{X}}, D_{\mathcal{Y}}, G_{\mathcal{X}}, G_{\mathcal{Y}},\right).
\end{aligned}
\end{equation*}

Overall, the corresponding BMO object for (bilevel) weighted image denoising and generation \cite{semibgan,jiang2023bilevel,bilevelgan} is
\begin{equation*} 
\begin{aligned} 
&\min_{x \in \mathcal{X}} \mathcal{R}_{m_1}(x,y,z):=\mathbb{E}_{\xi\in\mathcal{D}_{m_1}} f\left(x,y^*(x),z^*(x)\right), \\
\text{ s.t. } ~ &y^*(x) , z^*(x)\in  \underset{y \in \mathcal{Y}}{\arg\min} ~ \underset{z \in \mathcal{Z}}{\arg\max}~\mathbb{E}_{\zeta \in \mathcal{D}_{m_2}} g(x,y,z).
\end{aligned}
\end{equation*}

\begin{figure*}[!t]
\centering
\subfigure[Validation loss]{
\includegraphics[width=0.4\textwidth]{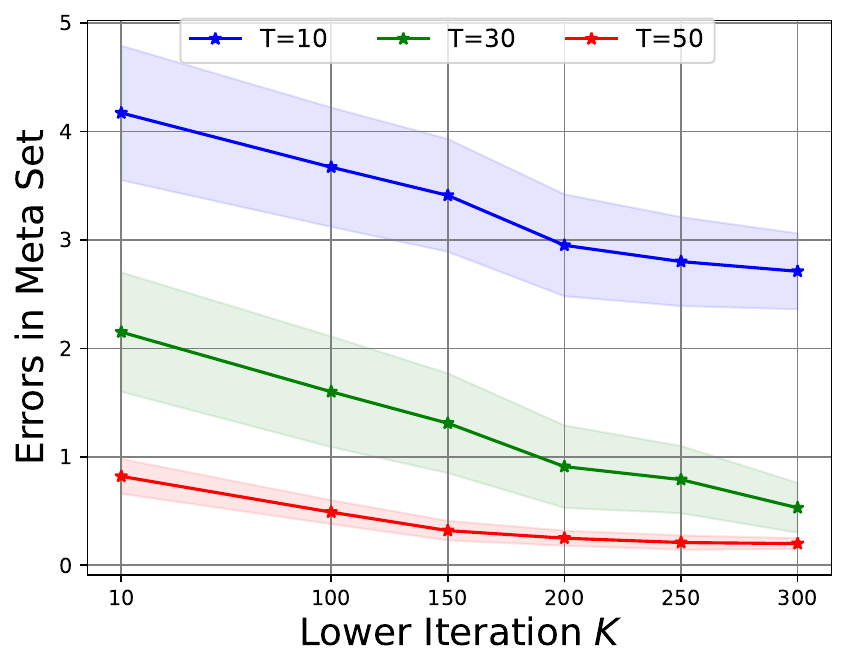}
}
\subfigure[Testing loss]{
\includegraphics[width=0.4\textwidth]{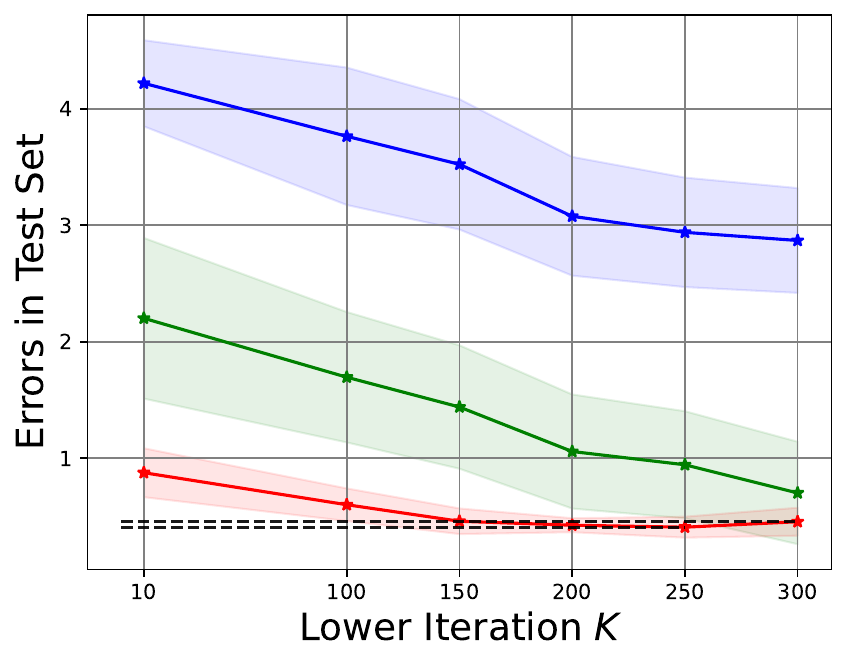}
}
\subfigure[Generalization Gap]{
\includegraphics[width=0.41\textwidth]{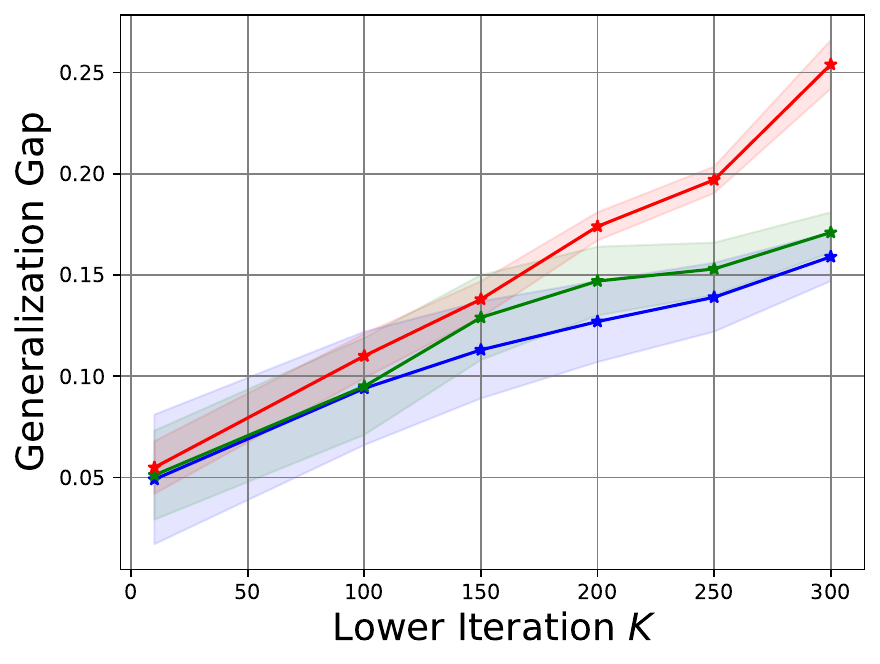}
}
\caption{Generalization validation with varying $T$ and $K$.}
\label{fig_gen1}
\end{figure*}

\subsection{Experimental Results}

Inspired by \citep{DBLP:conf/nips/BaoWLZZ21,zhang2024sbo}, we concentrate on assessing the generalization capability of the outer-level (upper-level) algorithm by evaluating its performance on the validation and testing datasets, where the generalization gap defined in \eqref{generalization_gap} is estimated by the differences between the validation error and the testing error.

\begin{figure*}[!t] 
\centering 
\centering
\subfigure[Validation loss]{
\includegraphics[width=0.4\textwidth]{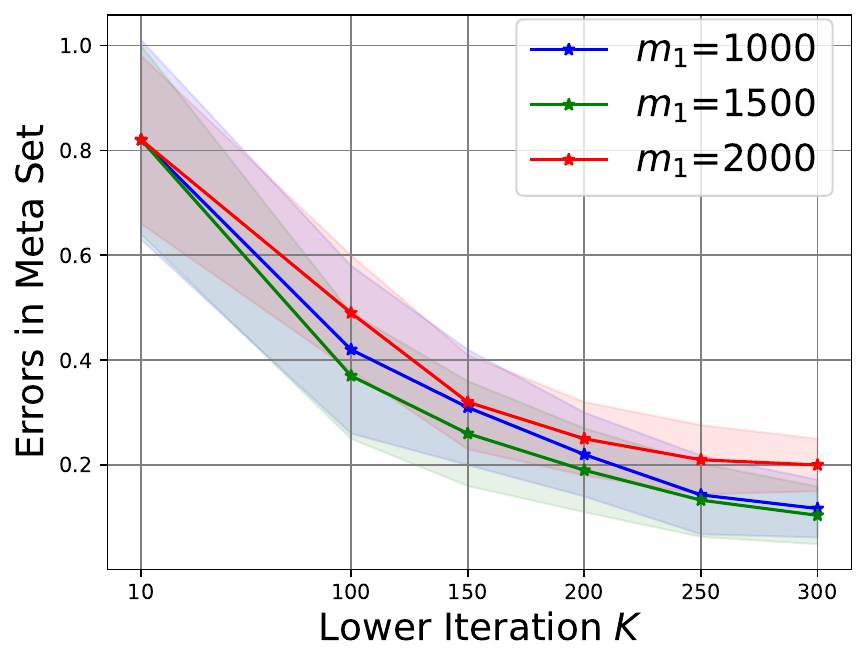}
}
\subfigure[Testing loss]{
\includegraphics[width=0.4\textwidth]{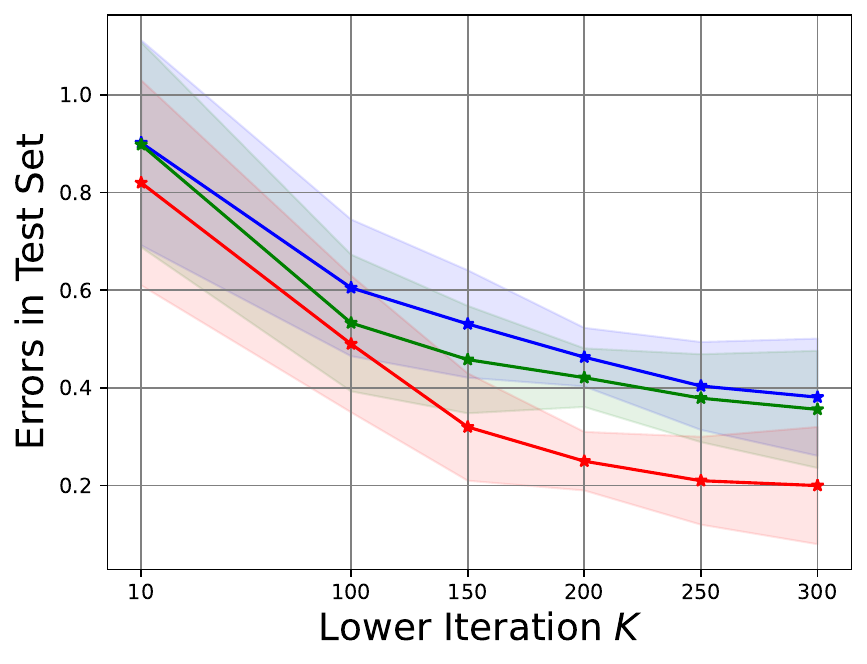}
}
\subfigure[Generalization Gap]{
\includegraphics[width=0.42\textwidth]{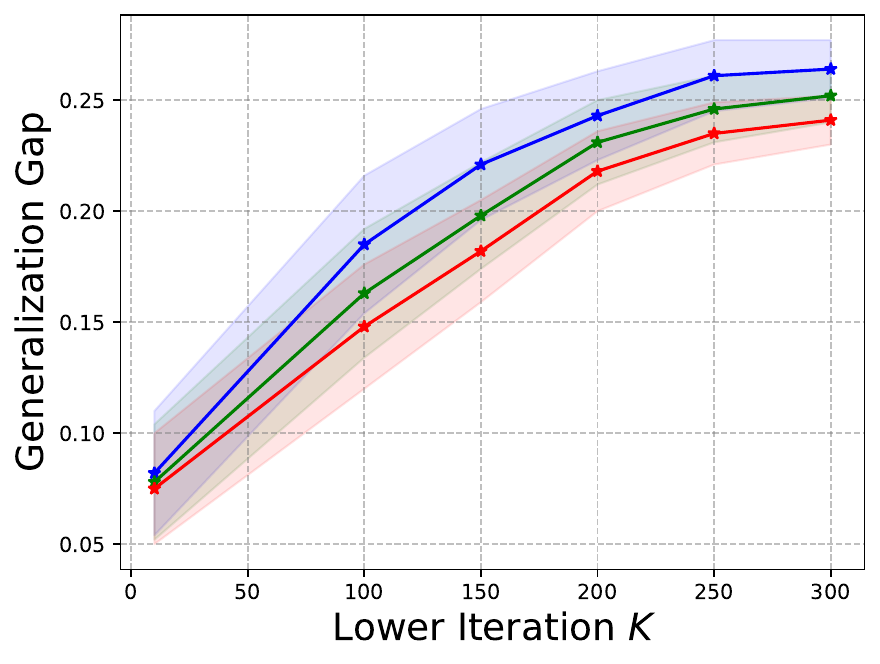}
}
\caption{Generalization validation with varying $m_1$ and $K$ (with fixed $T=50$).} 
\label{fig:figure2}
\end{figure*}

\begin{figure*}[!t] 
\centering
\includegraphics[width=0.55\linewidth]{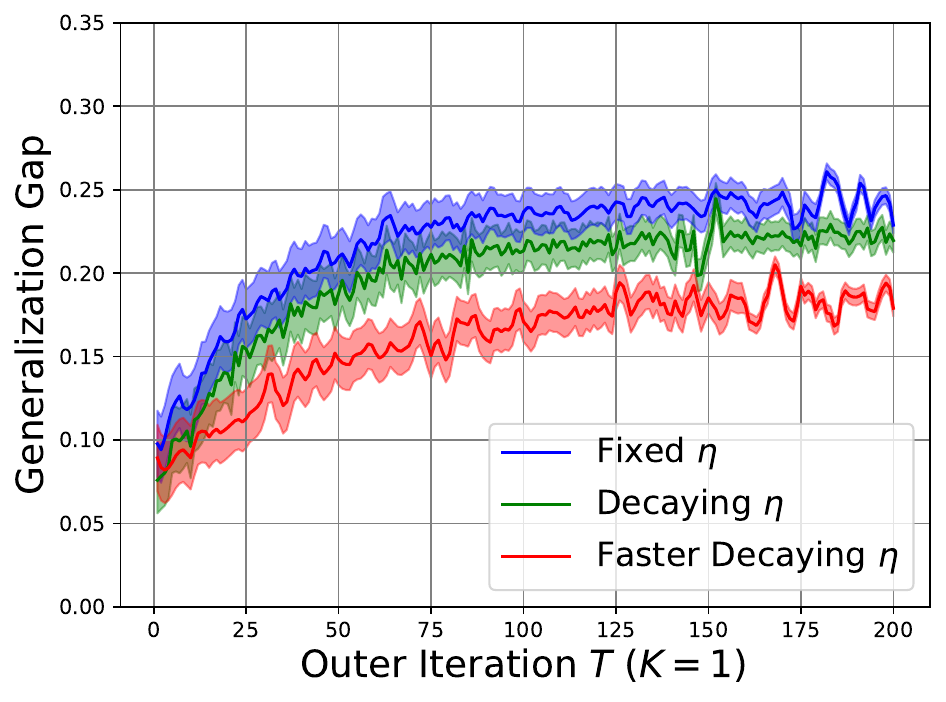} 
\caption{Generalization Gap with varying step size $\eta$ and $T$ ($K=1$).} 
\label{fig:figure3}
\end{figure*}

\begin{figure*}[!t]
\centering
\subfigure[Validation loss]{
\includegraphics[width=0.4\textwidth]{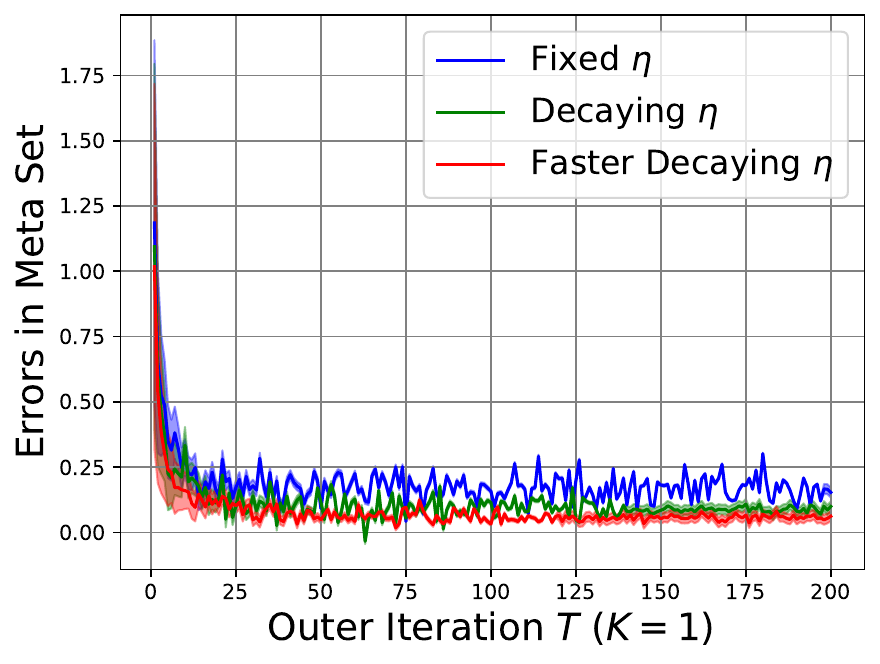}
}
\subfigure[Testing loss]{
\includegraphics[width=0.4\textwidth]{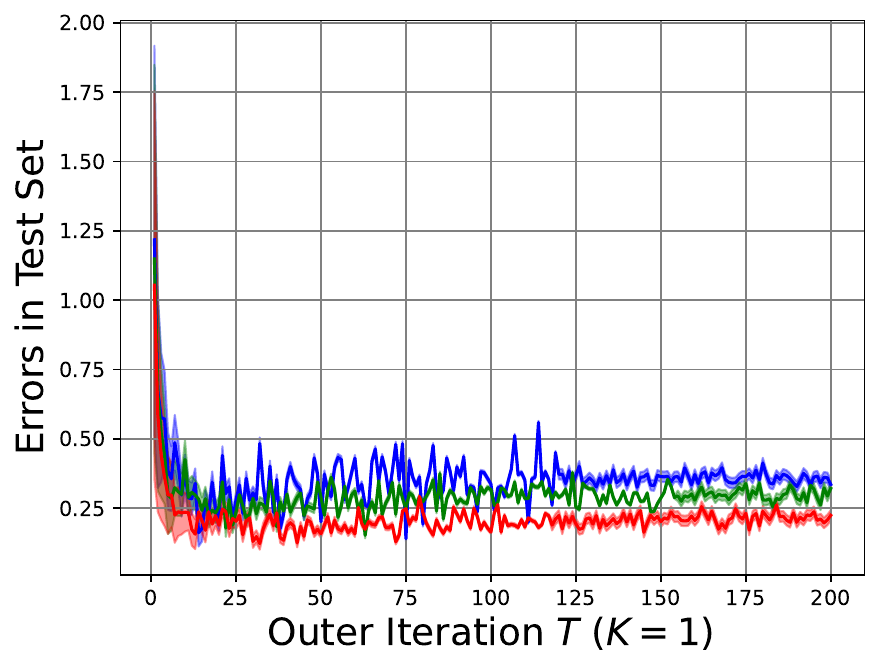}
}
\subfigure[Generalization Gap]{
\includegraphics[width=0.42\textwidth]{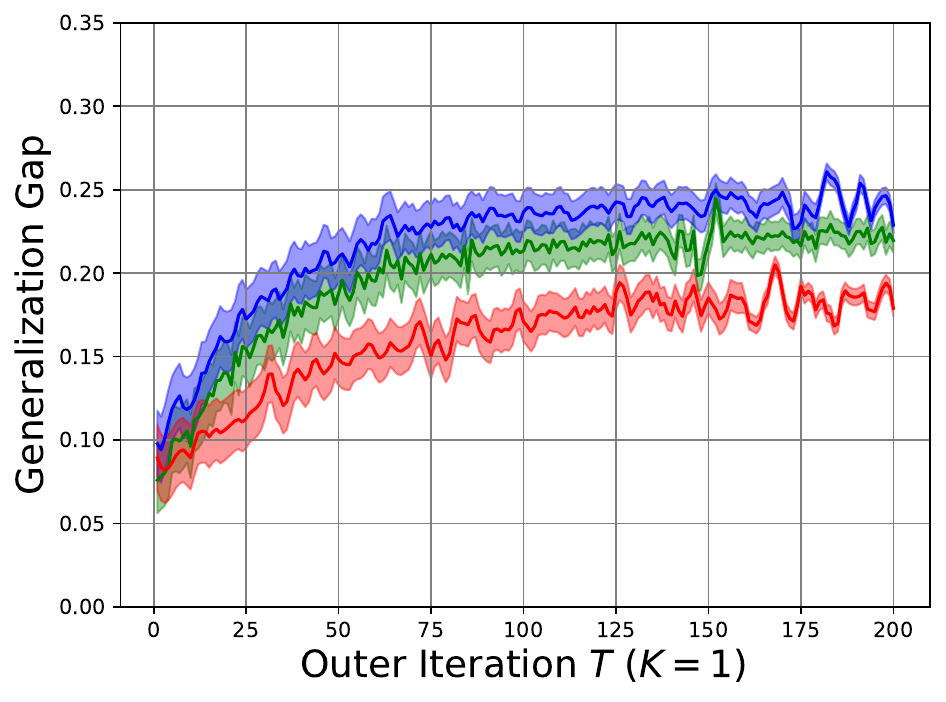}
}
\caption{Generalization validation with varying $\eta$ and $T$ (TSGDA-1 with $K=1$).}
\label{fig_stepsize}
\end{figure*}

\textbf{(a) Influence of iteration parameters $K$ and $T$ on generalization.} 
Figure \ref{fig_gen1} demonstrates that excessive outer ($K$) or inner ($T$) iterations may degrade generalization and bring under-fitting or overfitting in bilevel optimization, evidenced by increased testing error and generalization gap but decreased validation error. A practical trade-off exists between algorithmic convergence and generalization, as reflected in the setting of the maximum number of iterations \citep{DBLP:conf/nips/BaoWLZZ21}.

\textbf{(b) Influence of sample size $m_1$ and step sizes $\eta$ on generalization.}
Figure \ref{fig:figure2} reveals that slightly larger meta-set sizes ($m_1$) could accelerate convergence and enhance generalization. Figure \ref{fig:figure3} further establishes that step size ($\eta$) selection critically affects algorithmic generalization, consistent with our theoretical findings. Detailed results are left in \emph{Appendix A}.


\textbf{(c) Influence of varying step sizes $\eta$.}
In the following experiments, the step sizes are set to fixed values of $0.005$ and $10^{-3}$, with exponential decay rates of $0.95$ and $0.85$, respectively.

The result of Figure \ref{fig_stepsize} implies that step size ($\eta$) selection critically impacts generalization performance, where the error curve of faster-decayed step sizes reaches convergence faster and more stably. The empirical finding aligns with our theoretical conclusions.
In short, combining the theoretical analysis and empirical assessments from the main paper and the above extended experiments, provides insights into the generalization capabilities of the BMO. Specifically, by extension, some conclusions also practically apply to the broader class of bilevel optimization algorithms. Moreover, its generalization performance often improves with a larger meta set $m_1$ and judiciously chosen iteration counts $K$ and $T$. Insufficient iterations may lead to underfitting, while an excess can lead to overfitting. Typically, establishing appropriate learning rates is also instrumental for generalization.

The above empirical results largely verify our theoretical conclusions, indicating that generalization improves with larger $m_1$ but requires balanced $K$ and $T$. Insufficient iterations may lead to underfitting, while excessive iterations can induce overfitting. Optimal step sizes remain essential, and practical efficacy depends on calibrating sample sizes, step sizes, and the number of iterations.

\section{Conclusion} \label{section6}

This work establishes a systematic generalization analysis of BMO algorithms through the lens of on-average argument stability. We derive fine-grained bounds for SSGDA, TSGDA-1, and TSGDA-2 that explicitly characterize how generalization error scales with different practical settings. For SSGDA, we further bound its optimization error and excess risks, highlighting BMO’s intrinsic complexity. Deriving fine-grained lower bounds for BMO and MBO is left for our future work.

\bibliographystyle{elsarticle-num-names}
\bibliography{ref}

\end{document}